\documentclass{article}

\PassOptionsToPackage{round, comma, sort&compress, authoryear}{natbib}
\usepackage[preprint]{neurips_2025}
\usepackage[T1]{fontenc}
\usepackage[utf8]{inputenc}
\usepackage{microtype}
\usepackage{textcomp}

\PassOptionsToPackage{hyphens}{url}
\usepackage{url}
\usepackage{amsmath}
\usepackage{amssymb}

\usepackage{booktabs}
\usepackage{array}
\usepackage{longtable}
\usepackage{calc}
\providecommand{\real}[1]{#1}
\usepackage{enumitem}

\usepackage[font=small,labelfont=bf]{caption}

\usepackage{graphicx}
\graphicspath{{figures/}}

\usepackage{xcolor}

\usepackage[colorlinks=true,
            linkcolor=blue!50!black,
            citecolor=blue!50!black,
            urlcolor=blue!50!black]{hyperref}
\usepackage[capitalize, noabbrev]{cleveref}

\title{Output Dilution: Redundant but Fragile\\
  Representations in MoE Models}

\author{%
  Orion Reblitz-Richardson\thanks{%
    Distiller Labs. Correspondence to Distiller Labs
    \textless\texttt{orion@orionr.com}\textgreater.}
}

\date{August 2026}

\providecommand{\tightlist}{%
  \setlength{\itemsep}{0pt}\setlength{\parskip}{0pt}}

\usepackage{fancyvrb}
\DefineVerbatimEnvironment{Highlighting}{Verbatim}{commandchars=\\\{\}}
\newenvironment{Shaded}{\begin{quote}}{\end{quote}}
\newcommand{\KeywordTok}[1]{\textbf{#1}}

\newcommand{\StringTok}[1]{\textit{#1}}

\newcommand{\OperatorTok}[1]{#1}
\newcommand{\NormalTok}[1]{#1}
\newcommand{\BuiltInTok}[1]{#1}

\newcommand{\ControlFlowTok}[1]{\textbf{#1}}

\newcommand{\DecValTok}[1]{#1}

\begin{document}

\maketitle

\begin{abstract}
Mixture-of-Experts (MoE) models appear to encode moral content as
robustly as dense models, yet prove far more fragile in their encoding.
In OLMoE-1B-7B, linear probes recover moral valence from nearly every
expert-layer combination, with mean peak-layer accuracy above 90\%. But
these representations collapse under levels of activation noise that a
dense model of matched size easily tolerates, with a 4.2-fold
difference in robustness.

We trace this to \emph{output dilution}. Because the MoE block averages
across active experts before contributing to the residual stream, the
feedforward signal reaching downstream layers is nearly two orders of
magnitude smaller than in a dense MLP. Moral information, our interest,
survives aggregation intact but at a scale trivially overwhelmed by
perturbation. Routing itself remains stable under noise while the
vulnerability originates entirely in the diluted aggregate.

Checkpoint trajectories confirm this is architectural, not learned.
Experts never specialize and accuracy saturates within the first few
thousand steps. \emph{In sparse architectures, redundant encoding does
not imply robust encoding.}
\end{abstract}

\section{Introduction}\label{introduction}

Mixture-of-Experts (MoE) architectures route each token through a
sparse subset of expert modules, partitioning the representation
space into discrete, inspectable units. This structural partition
has a natural consequence for alignment research: if moral features
concentrate in specific experts, MoE models offer intervention
points (expert pruning, expert-specific fine-tuning, router
modification) that dense models lack. Conversely, if moral features
distribute uniformly across experts, MoE and dense architectures
are equivalent for alignment purposes, and the additional complexity
of expert-level analysis buys nothing.

We test this question on OLMoE-1B-7B \citep{muennighoff2024olmoe}, a
64-expert, top-8 MoE language model with 6.9B total parameters
(1.3B active per token), using the moral probing and fragility
methodology from companion work on dense OLMo models
\citep{reblitzrichardson2026fragility}. OLMoE is uniquely positioned for this
analysis: it is the open MoE model with by far the densest published
checkpoint record (244 checkpoints at 5,000-step intervals), and its
dense counterpart OLMo-2 1B (from the same
lab, with comparable active parameter count and full checkpoint
access) gives a controlled architectural comparison.

We report four findings that converge on a single mechanism:

\textbf{Finding 1: MoEs do not create expert moral specialization.} Nearly
all 1,024 per-expert probes (64 experts \(\times\) 16 layers) decode
moral content well above chance (1,020 exceed 75\%). At the peak layer, every expert
individually exceeds 84\% accuracy. The Gini coefficient of expert
accuracy is below 0.03 at all layers; moral encoding is as
uniformly distributed across experts as it is across neurons in a
dense model. The router shows negligible moral content preference
(maximum 1.8\%).

\textbf{Finding 2: MoE encoding is 4.2\(\times\) more fragile than dense.}
Despite matching dense OLMo-2 1B on probing accuracy (99.0\% vs.
99.0\% peak), OLMoE's moral encoding collapses under 4.2\(\times\)
less noise (mean critical \(\sigma^* = 0.92\) vs.~3.81). The
fragility gap is not explained by weaker individual expert
representations or unstable routing; both are robust in isolation.

\textbf{Finding 3: The fragility originates in output dilution.} The MoE
block's aggregated output (a top-8 weighted average of 64 expert
outputs) contributes to the residual stream at 74\(\times\) smaller
scale than the dense MLP output, measured as the standard deviation
of the feedforward block's output across inputs. This \emph{output
dilution} means that the same absolute noise level overwhelms the
MoE moral signal while leaving the dense signal intact.

\textbf{Finding 4: Specialization never emerges during training.} Across 11
checkpoints spanning OLMoE's training (step 5K to step 1.2M, covering
20B to 5,033B tokens), the Gini coefficient at the peak layer stays between
0.012 and 0.018 at every checkpoint (cf.~0.016--0.023 across all
layers in the final model, \S{}4.2). Moral encoding is present from
the earliest available checkpoint (92.1\% peak accuracy at step 5K)
and remains stable at 93.7\% by step 1.2M without ever concentrating
in specific experts. The top-5 experts by accuracy change between
adjacent checkpoints at near-random rates (Jaccard \(\approx\) 0.09).

The output dilution finding has direct implications for the
interpretability of probing accuracy as an alignment metric. Two
models can produce identical probing accuracy profiles (high
accuracy from early layers, broad encoding across the full network)
while differing by nearly two orders of magnitude in the robustness
of the underlying signal. Probing accuracy measures what information
is \emph{present}; fragility testing, as developed in companion work
\citep{reblitzrichardson2026fragility}, measures how \emph{securely} that information
is encoded. In MoE architectures, the gap between these two metrics
is dramatically larger than in dense models, because the sparse
aggregation bottleneck preserves information content while reducing
signal scale.

The paper contributes the first expert-level moral probing analysis
of an MoE language model, the first quantification of the MoE
output dilution effect and its relationship to representational
fragility, and a controlled dense-vs-MoE comparison on identical
probing methodology. All experiments run on a single MacBook Pro M4
Pro (24 GB, MPS) on base (non-instruct) models.

\section{Related Work}\label{related-work}

\textbf{Mixture-of-Experts architectures.} Sparse MoE was introduced by
\citet{shazeer2017moe} and scaled by \citet{fedus2022switch} and
\citet{lepikhin2021gshard}. Recent open MoE models include Mixtral
\citep{jiang2024mixtral}, DeepSeek-MoE \citep{dai2024deepseekmoe},
and OLMoE \citep{muennighoff2024olmoe}. OLMoE is unique in
publishing 244 training checkpoints, enabling trajectory analysis
unavailable for other MoE models.

\textbf{Expert specialization.} Prior work on what individual MoE experts
learn has focused on linguistic features (syntax, part-of-speech),
domain features (code vs.~natural language), and language-specific
specialization in multilingual models. \citet{zuo2022moe} find that
Switch Transformer experts partially specialize by token type.
\citet{chi2022expert} study expert utilization patterns. To our
knowledge, no prior work examines whether MoE experts specialize
for moral or ethical features.

\textbf{Moral probing in language models.} Probing classifiers
\citep{conneau2018probing,belinkov2022probing} train lightweight
classifiers on model-internal representations to test what
information is encoded. Moral probing specifically applies this
methodology to moral reasoning features, grounded in Moral
Foundations Theory \citep{haidt2012righteous,graham2013mft}.
Companion work \citep{reblitzrichardson2026fragility} develops the
layer-wise moral probing and fragility testing methodology we extend
to MoE models, establishing that fragility resolves structure after
probing accuracy saturates.

\textbf{Activation perturbation and representational robustness.}
Gaussian noise injection for probing robustness relates to work on
representation stability \citep{morcos2018representation} and
activation perturbation for identifying causally relevant features
\citep{vig2020causal,meng2022locating}. Our fragility protocol
\citep{reblitzrichardson2026fragility} adapts this approach to
alignment-relevant features, defining critical noise as a
quantitative robustness metric.

\textbf{Dense-model moral encoding.} The companion paper
\citep{reblitzrichardson2026fragility} establishes
that dense OLMo models encode moral features from early layers
(low encoding depth), broadly across the network (high encoding
breadth), with a fragility gradient that continues to resolve after
probing accuracy saturates. Prior work on this project also showed
that probe-direction suppression in dense 1B models does not
capture behavior due to feature redundancy, which motivated
investigating whether MoE's structural partition reduces this
redundancy.

\textbf{OLMo ecosystem.} OLMo \citep{groeneveld2024olmo} and OLMoE
\citep{muennighoff2024olmoe} are developed by the Allen Institute for
AI with a commitment to open science, including full training data,
code, intermediate checkpoints, and evaluation infrastructure. This
openness enables the controlled architectural comparison (\S{}4.1), the
output scale measurement (\S{}4.4), and the 11-checkpoint trajectory
analysis (\S{}4.5) that are central to our findings.

\section{Methodology}\label{methodology}

\subsection{Models and Comparison Design}\label{models-and-comparison-design}

\textbf{OLMoE-1B-7B} (\path|allenai/OLMoE-1B-7B-0924|; \citealp{muennighoff2024olmoe}) is a 16-layer MoE language model with 64 experts per layer,
top-8 routing, 6.9B total parameters (1.3B active per token), and
hidden dimension 2048. Each expert is a gated MLP with intermediate
dimension 1024, using SiLU activation. The router is a learned
linear projection (2048 \(\to\) 64) followed by softmax and top-\(k\)
selection with normalized weights. The model is trained with a
load-balancing auxiliary loss (\(\lambda = 0.01\)) to encourage
uniform expert utilization.

\textbf{OLMo-2 1B} (\path|allenai/OLMo-2-0425-1B|; \citealp{olmo2_2025})
is a 16-layer dense transformer with 1.5B parameters and hidden
dimension 2048. It serves as the architectural control: same lab,
same training philosophy, comparable active parameter count, same
number of layers and hidden dimension.

Both models are base (non-instruct) checkpoints. All experiments
use the same 240-pair moral probing dataset (\S{}3.5), the same probe
architecture (\S{}3.3), and the same fragility protocol (\S{}3.4).
Architecture is the independent variable.

\subsection{Per-Expert Activation Collection}\label{per-expert-activation-collection}

Standard layer-wise probing \citep{reblitzrichardson2026fragility} registers
forward hooks on transformer layer outputs to collect post-layer
hidden states. For per-expert probing, we bypass the router and
compute all 64 expert outputs in parallel.

For each input text, we hook \path|post_attention_layernorm| at each
layer to capture the pre-MoE hidden state \(h \in \mathbb{R}^{s
\times d}\) (where \(s\) is sequence length, \(d = 2048\)). We then
compute expert outputs by directly applying each expert's FFN
weights to the mean-pooled hidden state \(\bar{h} = \frac{1}{s}
\sum_t h_t\):

\[\text{gate\_up}_e = \bar{h} \cdot W^{\text{gate\_up}}_e{}^\top
\quad \in \mathbb{R}^{2k}\]

\[g_e, u_e = \text{chunk}(\text{gate\_up}_e) \quad \in
\mathbb{R}^k\]

\[o_e = \text{SiLU}(g_e) \odot u_e \cdot W^{\text{down}}_e{}^\top
\quad \in \mathbb{R}^d\]

where \(k = 1024\) is the intermediate dimension and \(e \in
\{0, \ldots, 63\}\). This computation is batched across all 64
experts using \path|torch.einsum|, yielding all expert outputs in a
single operation per layer.

For router analysis, we also capture the router logits by computing
\(\bar{h} \cdot W^{\text{gate}}{}^\top \in \mathbb{R}^{64}\), where
\(W^{\text{gate}}\) is the router's learned weight matrix.

\textbf{Clean aggregated output.} To produce the MoE block's actual
output for downstream probing, we apply the standard routing:
softmax over router logits, select top-8, normalize weights, and
compute the weighted sum of the selected experts' outputs.

\subsection{Probing Architecture}\label{probing-architecture}

All probes are binary linear classifiers: \texttt{nn.Linear(d,\ 1)} trained
with binary cross-entropy loss, Adam optimizer (lr \(= 10^{-2}\)), 50
epochs. For layer-level probes, \(d = 2048\) (full hidden dimension).
For per-expert probes, \(d = 2048\) (expert output dimension, which
equals hidden dimension in OLMoE's architecture). For aggregated-MoE
probes used in the perturbation experiments, \(d = 2048\).

The probe threshold is 0 (logit sign determines classification).
Accuracy is reported on a held-out test set (48 pairs, 96 texts).

\subsection{Fragility Protocol}\label{fragility-protocol}

We extend the fragility testing protocol from companion work
\citep{reblitzrichardson2026fragility}. In the standard protocol, Gaussian noise
\(\mathcal{N}(0, \sigma^2 I)\) is injected into post-layer hidden
states at magnitudes \(\sigma \in \{0.1, 0.3, 1.0, 3.0, 10.0\}\), with
accuracy averaged over 10 noise seeds per level, and the critical noise
\(\sigma^*\) is the smallest \(\sigma\) at which the seed-mean probe
accuracy drops below 0.6. Following the convention of
\citet{reblitzrichardson2026fragility}, a layer whose probe never drops
below threshold is censored at the grid maximum (not dropped) when
averaging, so \(\sigma^*\) aggregates are means over all layers.

For the MoE component perturbation experiment (\S{}4.4), we extend this
protocol to three perturbation targets within the MoE block:

\begin{enumerate}
\def\labelenumi{\arabic{enumi}.}
\tightlist
\item
  \textbf{Router perturbation.} Noise is added to the router logits
  before softmax and top-\(k\) selection. This changes both which
  experts are selected and their aggregation weights.
\item
  \textbf{Expert perturbation.} Noise is added to individual expert
  outputs before weighted aggregation. Routing is held fixed (clean
  logits determine expert selection and weights).
\item
  \textbf{Output perturbation.} Noise is added to the final aggregated
  MoE output (control condition equivalent to standard fragility
  testing).
\end{enumerate}

For each condition, probes are trained on clean aggregated outputs
(train set) and evaluated on perturbed outputs (test set) at each
noise level. Results are averaged over 10 random seeds per noise
level to reduce variance from individual noise realizations.

\textbf{Output scale measurement.} To interpret the component fragility
results, we measure the natural scale of each component (standard
deviation across test texts) and the feedforward output scale at each
layer for both OLMoE and OLMo-2 on the same input texts.

\subsection{Probing Dataset}\label{probing-dataset}

We use the same 240-pair moral probing dataset as companion work
\citep{reblitzrichardson2026fragility}: 40 minimal pairs per Moral
Foundations Theory foundation (care/harm, fairness/cheating,
loyalty/betrayal, authority/subversion, sanctity/degradation,
liberty/oppression), subsampled with a deterministic seed from a
1,200-pair dataset constructed per published quality guidelines
with LLM-assisted filtering for naturalness and moral neutrality
of neutral-side sentences (see \path|DATASET_GUIDELINES.md|). The
subsample is split 80/20 into 192 training pairs (384 texts) and
48 test pairs (96 texts), with foundation balance preserved across
splits.

Dataset identity is load-bearing: the dense-vs-MoE comparison (\S{}4.1)
and the output scale comparison (\S{}4.4) use identical inputs to ensure
any observed differences are architectural, not data-driven.

\subsection{Checkpoint Trajectory Analysis}\label{checkpoint-trajectory-analysis}

OLMoE publishes 244 training checkpoints at 5,000-step intervals
from step 5,000 (20B tokens) through step 1,220,000 (5,117B tokens).
We select 11 checkpoints spanning training: dense early sampling
(steps 5K, 10K, 20K, 50K, 100K) and logarithmic spacing through
the remainder (steps 200K, 400K, 600K, 800K, 1M, 1.2M). Our sample
therefore ends at step 1,200,000 (5,033B tokens), just short of the
published set's final step 1,220,000 (5,117B tokens); the 84B-token
gap does not affect any trajectory conclusion. At each
checkpoint, we run the full per-expert probing analysis (\S{}3.2--3.3)
and router analysis (\S{}3.2), computing the Gini coefficient of
per-expert moral accuracy and tracking expert identity stability
(Jaccard similarity of the top-5 experts between adjacent
checkpoints).

Each checkpoint is loaded sequentially (load, probe, free) to fit
within 24 GB memory. Results are saved per-checkpoint with resume
support, enabling interrupted runs to continue from the last
completed checkpoint.

\subsection{Hardware and Reproducibility}\label{hardware-and-reproducibility}

All experiments run on a MacBook Pro M4 Pro (24 GB unified memory)
using PyTorch MPS backend with float16 precision. OLMoE-1B-7B
requires \textasciitilde14 GB in float16; OLMo-2 1B requires \textasciitilde3 GB. Models are
loaded sequentially (load, evaluate, free) to fit within memory.

A monkey-patch to \path|torch.histc| is required for OLMoE on MPS: the
MoE router's token-counting operation uses integer \texttt{histc}, which
is not implemented on MPS or CPU. The patch casts to float and
falls back to CPU for this single operation.

All random seeds, model revisions, and command-line invocations are
recorded in the output JSON files. Experimental scripts are
available at \path|papers/2_moe_output_dilution/scripts/|.

\section{Results}\label{results}

\subsection{Dense vs.~MoE: Same Accuracy, Different Robustness}\label{dense-vs.-moe-same-accuracy-different-robustness}

We first establish the baseline comparison between OLMoE-1B-7B and
dense OLMo-2 1B using the standard layer-wise moral probing and
fragility battery from companion work \citep{reblitzrichardson2026fragility}.
Both models have 16 transformer layers and comparable active
parameter counts (1.3B active for OLMoE vs.~1.5B for OLMo-2),
enabling a controlled architectural comparison on the same 240-pair
probing dataset.

\textbf{Probing accuracy is indistinguishable.} OLMoE achieves peak
probing accuracy of 99.0\% at layer 13; OLMo-2 achieves 99.0\% at
layer 12. Both models reach onset (accuracy \(> 0.6\)) at layer 0 and
maintain encoding breadth of 1.0, meaning moral content is decodable from
every layer. The probing accuracy profiles differ only in that
OLMoE shows lower early-layer accuracy (79--86\% at layers 0--3 vs.
94--97\% for OLMo-2) before converging at later layers.

\textbf{Fragility diverges sharply.} Under Gaussian noise injection at
\(\sigma \in \{0.1, 0.3, 1.0, 3.0, 10.0\}\) averaged over 10 noise
seeds, OLMoE is 4.2\(\times\) more fragile than OLMo-2: mean critical
noise \(\sigma^* = 0.92\) vs.~3.81. The fragility profiles also differ
structurally. OLMo-2 shows distributed robustness, with critical
noise \(\geq\) 3.0 at 12 of 16 layers and \(\geq\) 10.0 at layers 13,
14, and 15. OLMoE concentrates robustness in the final three
layers only (critical noise 3.0 at layers 13--15; \(\leq\) 0.3 at
9 of 16 layers, and \(\leq\) 1.0 at 13 of 16).

\begin{figure}[t]
  \centering
  \includegraphics[width=\linewidth]{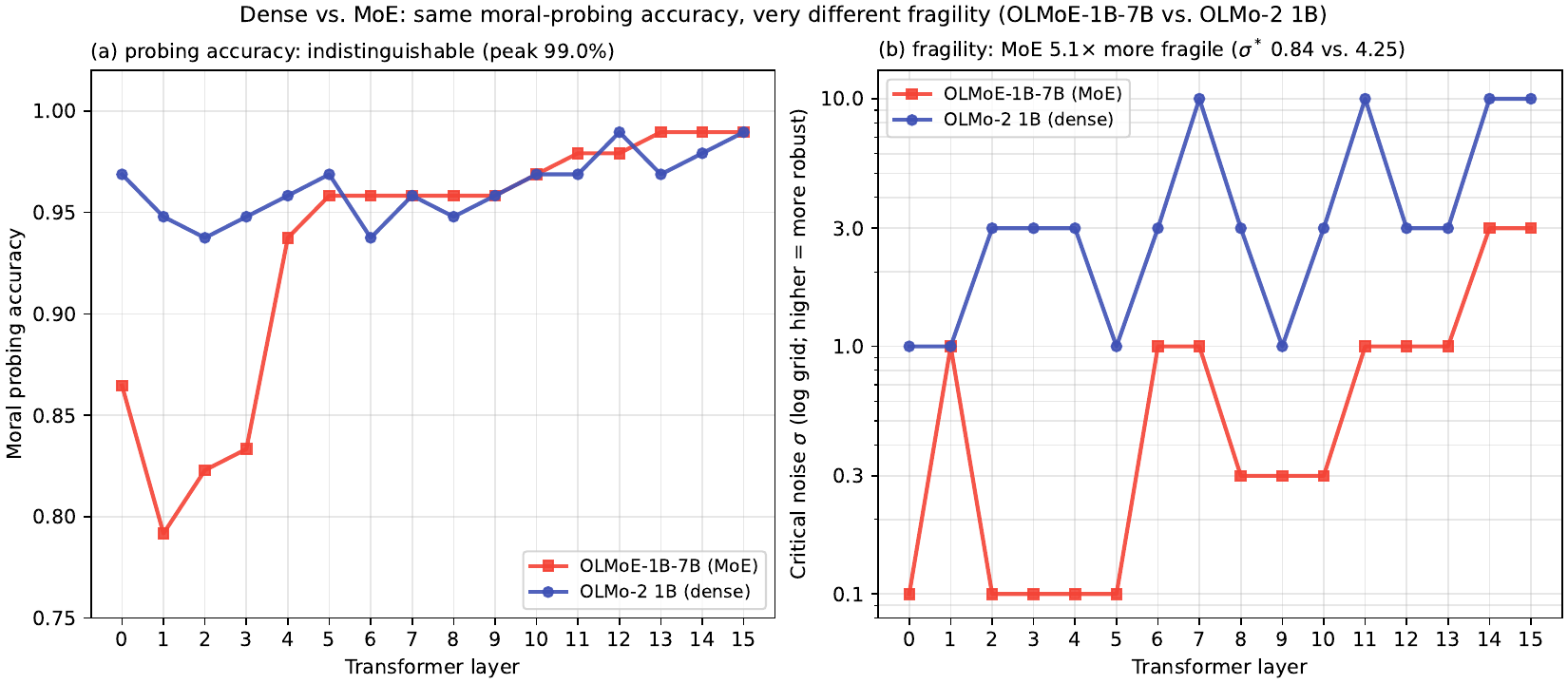}
  \caption{Dense and MoE encode moral content with near-identical accuracy but very different robustness. (a)~Per-layer moral probing accuracy for OLMoE-1B-7B and dense OLMo-2~1B; both peak at 99.0\%, differing only at the early layers. (b)~Per-layer critical noise $\sigma^*$ (smallest $\sigma$ at which probe accuracy falls below 0.6, on the log grid $\{0.1, 0.3, 1.0, 3.0, 10.0\}$): OLMoE is 5.1$\times$ more fragile (mean $\sigma^*$ 0.84 vs.\ 4.25) and concentrates robustness in the final two layers.}
  \label{fig:dense-vs-moe}
\end{figure}

\textbf{Figure~\ref{fig:dense-vs-moe}} contrasts the two architectures across both metrics.

This establishes the puzzle the remaining experiments investigate:
both architectures encode moral content with near-identical accuracy,
but the MoE encoding is substantially more fragile. What is it about
MoE that produces this gap?

\subsection{No Expert Moral Specialization}\label{no-expert-moral-specialization}

We trained 1,024 independent binary probes, one per expert-layer
combination (64 experts \(\times\) 16 layers), on per-expert
activations collected by bypassing the router and computing all 64
expert FFN outputs in parallel via batched einsum on the pre-MoE
hidden state. If MoE architectures create expert-level moral
specialization, we would expect a sparse subset of experts to achieve
high probe accuracy while most remain near chance.

\textbf{The result is the opposite: moral encoding is uniformly distributed
across all experts at every layer.} 1,020 of 1,024 expert probes
exceed 75\% accuracy (four exceptions at early layers 1--3, ranging
from 72--75\%). At the peak layer (layer 14),
all 64 experts individually exceed 84\% accuracy (mean 93.0\%, min
84.4\%). The per-layer Gini coefficient of expert accuracy, which measures
how concentrated moral signal is across experts, ranges from 0.016
to 0.023, indicating near-perfect uniformity. Gini is modestly
higher in early layers (0.021--0.023 at layers 0--3) and lowest in
mid-network (0.016 at layers 8, 9, and 12), suggesting that moral
encoding becomes \emph{more} uniform through the early and middle layers.

\begin{figure}[t]
  \centering
  \includegraphics[width=\linewidth]{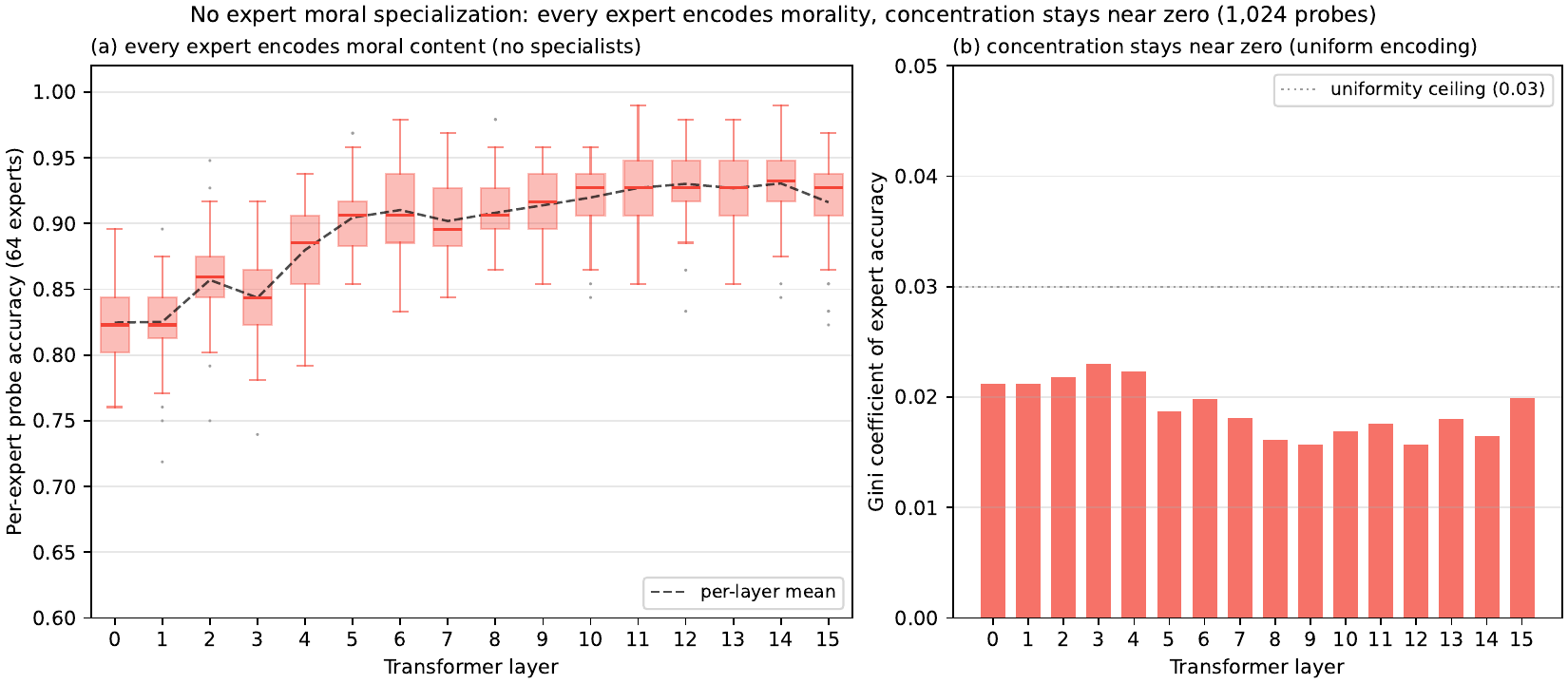}
  \caption{No expert moral specialization. (a)~Distribution of the 64 per-expert probe accuracies at each layer (box plot) with the per-layer mean overlaid; every expert encodes moral content, with no sparse high-accuracy subset. (b)~The per-layer Gini coefficient of expert accuracy stays in $[0.016, 0.023]$, far below any concentration threshold, and is lowest at the late layers where encoding peaks.}
  \label{fig:expert-uniformity}
\end{figure}

\textbf{Figure~\ref{fig:expert-uniformity}} shows the per-expert accuracy distribution at every layer.

This finding has immediate consequences for alignment interventions.
Dense models encode moral features diffusely across neurons within
each layer; MoE partitions representations across 64 discrete expert
modules, yet moral features remain equally diffuse across all 64.
The structural partition MoE introduces does not induce functional
specialization for moral content.

\subsection{Router Is Content-Agnostic for Morality}\label{router-is-content-agnostic-for-morality}

The absence of expert specialization raises the question of whether
the router treats moral and neutral inputs differently. We analyzed
per-layer routing distributions by comparing mean router probabilities
and top-8 selection frequencies conditioned on moral vs.~neutral
input texts.

\textbf{The router shows negligible moral preference.} The maximum routing
preference (the largest absolute difference in mean routing
probability between moral and neutral inputs for any single expert)
is 1.8\% (layer 12, expert 37). Using a threshold of 0.5\% absolute
routing-probability difference, the number of experts with any
detectable preference ranges from 3 (layer 1) to 19 (layer 6)
out of 64, but all preference magnitudes are small: the 95th
percentile across all 1,024 expert-layer combinations is below 2\%.

Combined with \S{}4.2, this establishes that moral encoding in OLMoE is
doubly diffuse: the router does not segregate moral tokens to
specific experts, and every expert that receives tokens encodes moral
content with comparable accuracy. MoE and dense architectures produce
equivalent moral encoding geometry despite their structural
differences.

\subsection{Output Dilution Explains MoE Fragility}\label{output-dilution-explains-moe-fragility}

Having established that moral encoding is uniformly distributed
across experts and that the router is content-agnostic, we turn to
the source of the 4.2\(\times\) fragility gap. We isolated three
perturbation targets within the MoE block:

\begin{itemize}
\tightlist
\item
  \textbf{Router perturbation}: Gaussian noise on router logits before
  softmax and top-\(k\) selection, changing which experts are selected
  and their aggregation weights.
\item
  \textbf{Expert perturbation}: Gaussian noise on individual expert
  outputs before weighted aggregation.
\item
  \textbf{Output perturbation}: Gaussian noise on the final aggregated
  MoE output (control condition matching \S{}4.1).
\end{itemize}

For each condition, probes were trained on clean aggregated MoE
outputs and evaluated on perturbed outputs at noise levels
\(\sigma \in \{0.01, 0.03, 0.1, 0.3, 1.0, 3.0, 10.0\}\), averaged
over 10 random seeds.

\textbf{The component fragility ranking reverses the natural hypothesis.}
The router is the \emph{most robust} component: 8 of 16 layers never reach
the fragility threshold at any tested noise level. Following the
cap-at-maximum convention (\S{}3.4) those never-fragile layers are
censored at \(\sigma=10\) rather than dropped, giving mean critical
noise \(\sigma^* = 9.56\) (it is 9.13 if they are dropped instead).
Expert outputs are moderately fragile (\(\sigma^* = 1.8\), all 16
layers fragile). The aggregated output is the most fragile
(\(\sigma^* = 0.6\), all 16 layers fragile), consistent with the
full-hidden-state fragility from \S{}4.1.

This counterintuitive ranking is explained by the natural scales of
each component. The MoE block's aggregated output has a standard
deviation of only 0.003--0.008 at layers 0--8, orders of magnitude
smaller than the router logit scale (\textasciitilde0.5) and comparable to the
smallest tested noise levels.

\subsubsection{\texorpdfstring{The 74\(\times\) output scale gap}{The 74\textbackslash times output scale gap}}\label{the-74times-output-scale-gap}

To test whether this small output scale is an inherent property of
MoE aggregation, we directly measured the feedforward output scale at
every layer for both OLMoE and OLMo-2 on the same 100 input texts.
The dense MLP produces outputs \textbf{74\(\times\) larger on average} than
the MoE block, measured as the standard deviation of the mean-pooled
feedforward output across texts:

{\def\LTcaptype{none} 
\begin{longtable}[]{@{}rccc@{}}
\toprule\noalign{}
Layer & OLMoE MoE std & OLMo MLP std & Ratio \\
\midrule\noalign{}
\endhead
\bottomrule\noalign{}
\endlastfoot
0 & 0.003 & 0.448 & 167\(\times\) \\
5 & 0.003 & 0.291 & 108\(\times\) \\
8 & 0.008 & 0.459 & 60\(\times\) \\
12 & 0.018 & 1.070 & 61\(\times\) \\
15 & 0.096 & 8.779 & 91\(\times\) \\
\end{longtable}
}

The ratio exceeds 60\(\times\) at 9 of 16 layers. The per-layer ratio
varies widely, from 5.3\(\times\) (layer 2, where the MoE output scale
spikes) to 167\(\times\) (layer 0); the table above excerpts five
representative layers. The MoE block's contribution to the residual
stream is not just smaller; it operates on a fundamentally different
scale than the dense MLP.

\begin{figure}[t]
  \centering
  \includegraphics[width=\linewidth]{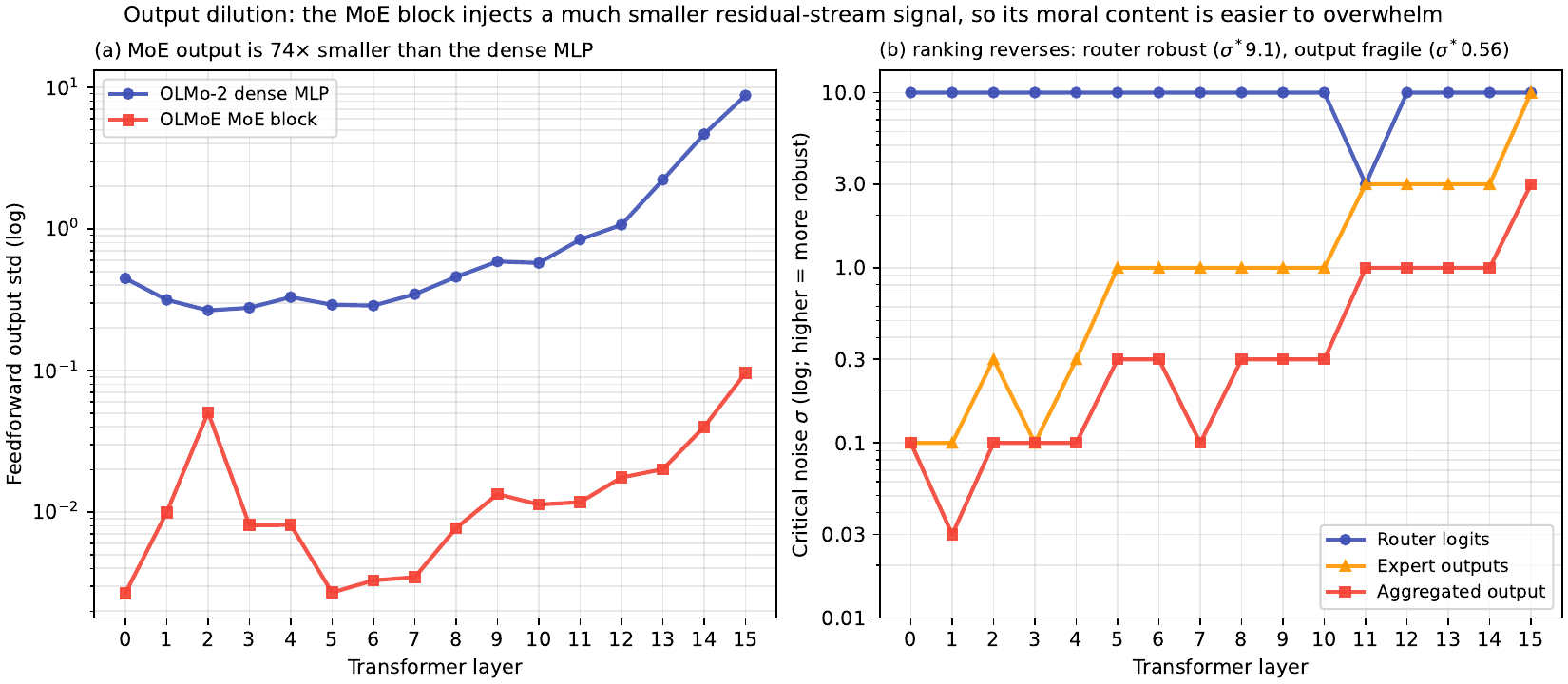}
  \caption{Output dilution explains MoE fragility. (a)~Per-layer feedforward output scale (standard deviation of the mean-pooled output) for the OLMoE MoE block vs.\ the dense OLMo-2 MLP; the dense MLP output is 74$\times$ larger on average. (b)~Per-layer critical noise for the three MoE perturbation targets: the router is most robust (mean $\sigma^*$ 9.1), the aggregated output most fragile (mean $\sigma^*$ 0.56), because the output operates on the diluted scale from panel~(a).}
  \label{fig:output-dilution}
\end{figure}

\textbf{Figure~\ref{fig:output-dilution}} relates the output-scale gap to component fragility.

\textbf{This output dilution is the mechanism behind MoE fragility.}
Because only 8 of 64 experts contribute to each token's MoE output,
and the routing weights further attenuate each expert's contribution,
the MoE block injects a much smaller perturbation into the residual
stream than a dense MLP. The moral signal carried by this small
perturbation is correspondingly easier to overwhelm with noise. This
is precisely raw critical noise behaving as a \emph{scale meter}, and the
companion dense-model study \citep{reblitzrichardson2026fragility}
independently establishes the same coupling at the within-model level:
an RMS-normalized control there shows that per-layer activation scale
drives raw fragility, so a fixed-architecture comparison like ours,
where the \(74\times\) output-scale gap maps onto a \(4.2\times\) fragility
gap, is reading scale exactly as intended.

The finding cleanly connects all four prior results:

\begin{enumerate}
\def\labelenumi{\arabic{enumi}.}
\tightlist
\item
  \textbf{Probing accuracy is preserved} (\S{}4.1) because the MoE output,
  though small, contains the same information content as the dense
  MLP output; a linear probe with learned weights can amplify
  the signal.
\item
  \textbf{No expert specialization} (\S{}4.2) because every expert processes
  the same pre-MoE hidden state and applies the same architectural
  pattern; specialization would require the router to route moral
  content selectively, which it does not (\S{}4.3).
\item
  \textbf{Fragility increases} (\S{}4.1) because the absolute noise
  threshold to disrupt a 0.003-scale signal is much lower than for
  a 0.3-scale signal.
\item
  \textbf{Router robustness} (\S{}4.4) because the routing mechanism
  operates on logits at scale \textasciitilde0.5, far above the noise levels
  that disrupt the MoE output.
\end{enumerate}

\subsection{Specialization Never Emerges During Training}\label{specialization-never-emerges-during-training}

OLMoE publishes 244 training checkpoints at 5,000-step intervals,
spanning from step 5,000 (20B tokens) to step 1,220,000 (5,117B
tokens). We ran the per-expert probing analysis (\S{}4.2) and router
analysis (\S{}4.3) at 11 checkpoints spanning training: dense early
sampling (steps 5K, 10K, 20K, 50K, 100K) and logarithmic spacing
through the remainder (steps 200K, 400K, 600K, 800K, 1M, 1.2M).

\textbf{Moral encoding appears from the earliest available checkpoint.}
At step 5,000 (20B tokens, \textasciitilde0.4\% of training), per-expert mean
accuracy already reaches 92.1\% at the peak layer, with 1,006 of
1,024 expert probes above 75\%. Accuracy is remarkably stable
throughout training (93.6\% at step 10K, 93.2\% at step 200K,
93.7\% at step 1.2M), fluctuating in a narrow 92--94\% band rather
than progressively sharpening. The peak layer stabilizes at layer
14 from step 200K onward, matching the final model's peak.

\textbf{Specialization never appears at any checkpoint.} The Gini
coefficient of per-expert accuracy remains between 0.012 and
0.018 at the peak layer across all 11 checkpoints, never exceeding
0.03 at any layer of any checkpoint. The trajectory plot shows
accuracy stable while Gini stays flat: the model maintains uniform
moral representations throughout training without concentrating
them in specific experts. Overall mean Gini (averaged across all 16
layers) shows a mild \emph{decrease} from 0.020 at step 50K to 0.018
at step 1M, suggesting that training produces more \emph{uniform}
encoding, not more specialized.

\begin{figure}[t]
  \centering
  \includegraphics[width=\linewidth]{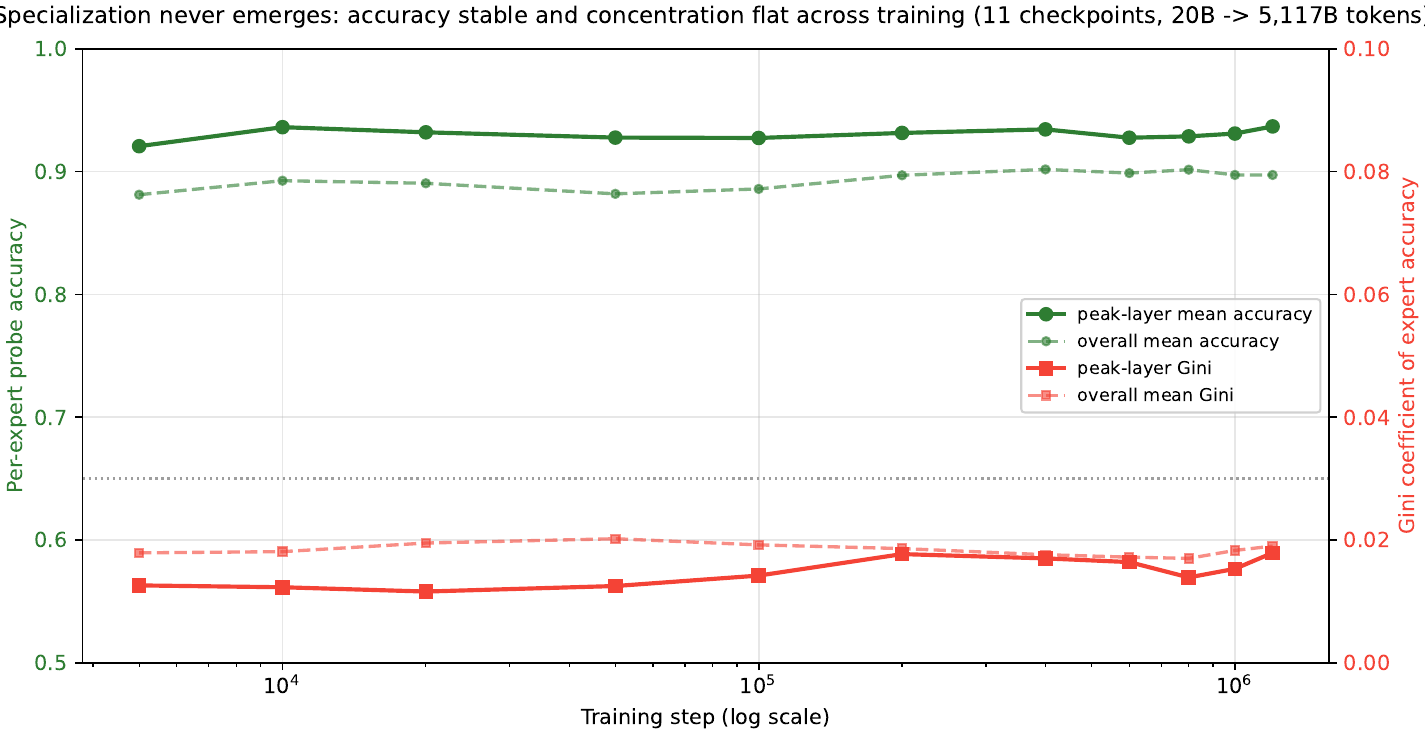}
  \caption{Specialization never emerges during training. Across 11 OLMoE checkpoints (step 5K--1.2M, 20B--5{,}117B tokens), peak-layer and overall mean per-expert accuracy (left axis) stay in a 92--94\% band from the earliest checkpoint, while the Gini coefficient of expert accuracy (right axis) stays flat near zero. Moral encoding is present from the start and never concentrates into specific experts.}
  \label{fig:training-trajectory}
\end{figure}

\textbf{Figure~\ref{fig:training-trajectory}} plots the training trajectory of accuracy and concentration.

\textbf{Expert identity is unstable.} The Jaccard similarity of the
top-5 highest-accuracy experts between adjacent checkpoints
fluctuates near the random baseline of \(5/64 \approx 0.08\),
ranging from 0.0 (complete turnover) to 0.25 (two shared experts
out of five). No stable ``moral expert'' identity exists; the
ranking of experts by moral accuracy is noise around a uniform
mean, not a consistent specialization pattern.

\section{Discussion}\label{discussion}

\subsection{Output Dilution as an Architectural Property}\label{output-dilution-as-an-architectural-property}

The 74\(\times\) output scale gap between MoE and dense feedforward
blocks is not specific to moral encoding; it is a structural
consequence of sparse expert aggregation. When a top-\(k\) routing
mechanism selects 8 of 64 experts, each expert contributes roughly
\(\frac{1}{8}\) of the aggregated output (modulated by routing
weights). The aggregated output is therefore a weighted average of 8
expert outputs, each operating on a 1024-dimensional intermediate
space, producing a 2048-dimensional output. The dense MLP, by
contrast, applies its full parameter budget to every token, producing
a larger-scale output.

The dilution effect likely scales with the sparsity ratio. OLMoE
uses top-8 of 64 (12.5\% sparsity); models with higher sparsity
(e.g., top-2 of 8 in Mixtral) may show even stronger dilution,
while lower sparsity would reduce it. The load-balancing auxiliary
loss, which encourages uniform expert utilization, may further
contribute by preventing any single expert from dominating the
aggregated output.

An important caveat: the output dilution we measure is the
\emph{feedforward block's contribution to the residual stream}, not the
total signal in the residual stream. Because the residual connection
carries forward the pre-MoE hidden state, the full hidden state
after the MoE block is dominated by the residual, not the MoE output.
This means that probing the full hidden state (as in \S{}4.1) picks up
moral signal from both the MoE contribution and accumulated residual
contributions from earlier layers. The fragility difference arises
because noise added to the full hidden state disrupts the MoE
contribution disproportionately: the noise is small relative to the
residual but large relative to the MoE output.

\subsection{Implications for Alignment Interventions}\label{implications-for-alignment-interventions}

The absence of expert moral specialization (\S{}4.2) closes one
potential intervention path: there are no ``moral experts'' to prune,
fine-tune, or monitor. MoE architectures, despite their structural
partition into discrete expert modules, do not make moral encoding
more tractable for targeted intervention than dense models do.

The output dilution finding (\S{}4.4), however, opens a different
concern. If moral features in MoE models are encoded at very small
scale in the feedforward contribution, they may be easier to
accidentally destroy during fine-tuning. A LoRA adapter that
modifies the feedforward weights by even a small amount in absolute
terms could represent a large \emph{relative} perturbation to the MoE
output. This prediction is testable: companion work's C15 finding
(fragility-locus shift under insecure-code LoRA in dense OLMo)
could be replicated on OLMoE, where we would predict a larger
fragility shift from the same fine-tuning recipe.

\subsection{Feature Redundancy Across Architectures}\label{feature-redundancy-across-architectures}

Prior work on this project \citep{reblitzrichardson2026fragility} found that
probe-direction suppression in dense 1B models does not capture
behavior: a gradient penalty suppresses the probe direction by
3.07 SD with no effect on behavioral judge scores (within 0.01 / 10).
This was attributed to feature redundancy: at the 1B scale, the
model has enough representational capacity to encode persona features
along directions orthogonal to the probe's extracted direction.

The present findings show that MoE architecture does not resolve
this redundancy problem. Despite partitioning representations across
64 discrete modules, moral features are encoded equally strongly
in every module. The structural partition of MoE is orthogonal to
the functional organization of moral features; the model encodes
the same information in every expert, just as a dense model encodes
it across every neuron.

The checkpoint trajectory analysis (\S{}4.5) strengthens this
conclusion: the absence of specialization is not a late-training
convergence but is present from step 5K (20B tokens), before the
routing mechanism has fully matured. Training does not create and
then destroy expert moral specialization; it never exists.

Feature redundancy in language models is therefore not a
consequence of architectural homogeneity (all neurons participating
in everything) but of training dynamics: the training objective
distributes useful features across all available representational
capacity, regardless of how that capacity is architecturally
partitioned. Load-balancing losses in MoE, which encourage uniform
expert utilization, may actively reinforce this tendency.

\subsection{Probing Accuracy as an Alignment Metric}\label{probing-accuracy-as-an-alignment-metric}

The dense-vs-MoE comparison starkly illustrates the insufficiency
of probing accuracy as a standalone alignment metric. Both models
achieve near-perfect probing accuracy (99--100\%) with full encoding
breadth (decodable at every layer) and near-zero encoding depth
(onset at layer 0). On probing accuracy alone, the two architectures
are indistinguishable. Yet the MoE model's moral encoding is
4.2\(\times\) more fragile, and the underlying feedforward signal is
74\(\times\) weaker.

Probing accuracy measures the \emph{presence} of information: whether
a linear classifier can extract a feature from the representation.
Fragility testing measures the \emph{security} of that information:
how much perturbation the encoding can withstand before the feature
becomes unextractable. The output dilution mechanism shows that
these two metrics can diverge dramatically: information can be
present (high accuracy) but insecure (low fragility), encoded at a
scale that is trivially disrupted.

The early-layer accuracy gap between architectures offers a
second diagnostic. OLMoE's early layers (0--3) achieve only 79--86\%
per-expert accuracy, compared to 94--97\% for OLMo-2 at the same
layers. When the probing dataset was tightened to remove
superficial cues (animacy and register confounds; see \S{}3.1 of
\citet{reblitzrichardson2026fragility}), OLMoE's early-layer
accuracy dropped more than OLMo-2's, suggesting that the MoE
architecture's diluted output makes early layers more dependent on
shallow features. This is consistent with output dilution: when the
feedforward contribution to the residual stream is small, early
layers cannot inject enough signal to support robust classification,
and probes compensate by exploiting dataset artifacts when available.

This reinforces the methodological argument from companion work
that fragility testing is a necessary complement to probing
accuracy, particularly when comparing architectures with different
internal signal scales.

\subsection{Limitations}\label{limitations}

\textbf{Single MoE model family.} We study only OLMoE. Generalization
to Mixtral (top-2 of 8), DeepSeek-MoE (fine-grained experts), or
Qwen-MoE is open. The output dilution mechanism predicts that
higher-sparsity architectures (lower \(k/N\) ratio) will show
greater fragility, but this has not been tested.

\textbf{Mean-pooling approximation.} Per-expert probing and the
perturbation experiments operate on mean-pooled representations,
collapsing the sequence dimension. This approximation is standard
in probing studies but may mask per-token routing effects. The
router's actual operation is per-token, not per-sequence.

\textbf{Linear probes only.} As in companion work, all probes are linear.
Nonlinear probes (MLP classifiers) might extract moral features from
the small-scale MoE output more effectively, potentially reducing
the apparent fragility gap. However, the output scale measurement
(\S{}4.4) is independent of probe architecture.

\textbf{Controlled but not identical comparison.} OLMoE and OLMo-2 are
from the same lab but differ in training data mix, hyperparameters,
and training duration, not just architecture. Same-lab provenance
minimizes but does not eliminate these confounds.

\textbf{English and MFT only.} The probing dataset covers English
sentences grounded in Haidt's Moral Foundations Theory. Moral
encoding in other languages, moral frameworks, or culturally
specific ethical norms is not tested.

\section{Conclusion}\label{conclusion}

We asked whether Mixture-of-Experts architectures create expert-level
moral specialization, discrete modules that concentrate moral
features and offer natural intervention points for alignment. The
answer is no. In OLMoE-1B-7B, all 64 experts at every layer encode
moral content with comparable accuracy (Gini \(< 0.03\)), and the
router shows no preference for routing moral content to specific
experts (maximum preference 1.8\%).

This null on specialization led to a positive finding about
architecture. MoE models are 4.2\(\times\) more fragile than dense
models on moral probing despite matching on accuracy, and the
mechanism is \emph{output dilution}: the MoE block's contribution to the
residual stream is 74\(\times\) smaller in scale than the dense MLP's,
because sparse aggregation (top-8 of 64 experts) attenuates each
expert's contribution. The moral signal is present but encoded at a
scale that is trivially overwhelmed by noise.

The checkpoint trajectory analysis (\S{}4.5) deepens the null: the
absence of specialization is not a late-training convergence but is
present from the earliest available checkpoint (step 5K, 20B tokens).
Moral encoding appears before the routing mechanism has fully matured,
and the Gini coefficient of per-expert accuracy remains below 0.03
throughout training. Training does not create or destroy expert moral
specialization; it was never there.

This finding refines the methodological program of companion work
\citep{reblitzrichardson2026fragility}. That work established fragility testing
as a complement to probing accuracy for tracking alignment depth
during pre-training. The present work shows that the gap between
probing accuracy and fragility is not just a temporal phenomenon
(fragility resolving after accuracy saturates) but an architectural
one: MoE's sparse aggregation creates a permanent structural
fragility that no amount of training can resolve without changing
the aggregation mechanism.

For future work, the output dilution mechanism makes specific
predictions. Models with higher sparsity (lower \(k/N\) ratio) should
show greater fragility. Fine-tuning should produce larger fragility
shifts in MoE than in dense models of comparable active size. And
MoE architectures that aggregate expert outputs differently
(concatenation, attention-based mixing, or denser routing) should
show correspondingly different fragility profiles.

\begin{ack}
This work made extensive use of Anthropic's Claude (the Claude Code agent on
Opus~4.6, 4.7, 4.8 and Fable~5) for code scaffolding, experimental scripts, and
prose drafting. The author retains responsibility for experimental design, all
scientific claims, and final wording.
\end{ack}

\bibliography{references}

\appendix
\newpage
\begin{center}
  \rule{0.5\linewidth}{0.4pt}\\[0.6em]
  {\Large\bfseries Appendices}\\[0.25em]
  {\small Supplementary material. Sections referenced from the main
   paper as ``Appendix A''--``Appendix D''.}\\[0.4em]
  \rule{0.5\linewidth}{0.4pt}
\end{center}
\vspace{0.5em}

\section{Probing dataset construction}\label{appendix-a.-probing-dataset-construction}

The 240-pair moral probing dataset used throughout this paper is
a deterministic subsample of a 1,200-pair dataset constructed per
published quality guidelines (\path|DATASET_GUIDELINES.md|) with
LLM-assisted filtering. The pipeline and full dataset are described
in companion work \citep{reblitzrichardson2026fragility}; we summarize
the construction here for self-containment.

\subsection{Seed extraction and pair generation}\label{a.1-seed-extraction-and-pair-generation}

The 1,200-pair parent dataset covers six Moral Foundations Theory
foundations \citep{haidt2012righteous,graham2013mft}: care/harm,
fairness/cheating, loyalty/betrayal, authority/subversion,
sanctity/degradation, and liberty/oppression (200 pairs per
foundation). Each pair consists of a moral sentence and a matched
neutral sentence that preserves syntactic structure and topic domain
while removing moral content. For example:

{\def\LTcaptype{none} 
\begin{longtable}[]{@{}
  >{\raggedright\arraybackslash}p{(\linewidth - 4\tabcolsep) * \real{0.3333}}
  >{\raggedright\arraybackslash}p{(\linewidth - 4\tabcolsep) * \real{0.3333}}
  >{\raggedright\arraybackslash}p{(\linewidth - 4\tabcolsep) * \real{0.3333}}@{}}
\toprule\noalign{}
\begin{minipage}[b]{\linewidth}\raggedright
Foundation
\end{minipage} & \begin{minipage}[b]{\linewidth}\raggedright
Moral
\end{minipage} & \begin{minipage}[b]{\linewidth}\raggedright
Neutral
\end{minipage} \\
\midrule\noalign{}
\endhead
\bottomrule\noalign{}
\endlastfoot
care/harm & ``Offering shelter to someone stranded in a storm, everyone helped.'' & ``Offering directions to someone lost in a new city, everyone helped.'' \\
fairness & ``The manager promoted the most qualified candidate despite personal ties.'' & ``The manager promoted the candidate who had applied first.'' \\
loyalty & ``She reported her company's illegal dumping to protect the community.'' & ``She reported her company's quarterly earnings to the board.'' \\
\end{longtable}
}

Neutral sentences are generated with LLM assistance and filtered
for naturalness and moral neutrality of the neutral side.

\subsection{Automated validation gates}\label{a.2-automated-validation-gates}

Pairs pass length-ratio gates (\(\leq\) 1.5 ratio), keyword filtering
(no explicit moral keywords in neutral sentences), and
deduplication. The 1,200-pair dataset is released alongside the
companion paper.

\subsection{Subsampling}\label{a.3-subsampling}

240 pairs (40 per foundation) are subsampled from the 1,200-pair
parent with a deterministic seed (42). The subsample is split
80/20 into 192 training pairs (384 texts) and 48 test pairs (96
texts), with foundation balance preserved across splits.

\subsection{Dataset identity across experiments}\label{a.4-dataset-identity-across-experiments}

All experiments in this paper use the identical 240-pair subsample.
The dense-vs-MoE comparison (\S 4.1), per-expert probing (\S 4.2),
routing analysis (\S 4.3), and output scale comparison (\S 4.4) all
process the same input texts, ensuring any observed differences are
architectural rather than data-driven.

\section{Per-expert probing details}\label{appendix-b.-per-expert-probing-details}

\subsection{Full accuracy statistics by layer}\label{b.1-full-accuracy-statistics-by-layer}

The following table reports the full per-expert probe
accuracy distribution across all 16 layers. ``Above 90\%'' counts
experts whose binary moral probe exceeds 90\% accuracy on the 96-text
test set; ``Below 60\%'' counts experts near chance.

{\def\LTcaptype{none} 
\begin{longtable}[]{@{}rrrrrrrr@{}}
\toprule\noalign{}
Layer & Mean & Std & Min & Max & Gini & \textgreater90\% & \textless60\% \\
\midrule\noalign{}
\endhead
\bottomrule\noalign{}
\endlastfoot
0 & 0.825 & 0.031 & 0.760 & 0.896 & 0.021 & 0 & 0 \\
1 & 0.825 & 0.032 & 0.719 & 0.896 & 0.021 & 0 & 0 \\
2 & 0.857 & 0.034 & 0.750 & 0.948 & 0.022 & 7 & 0 \\
3 & 0.844 & 0.035 & 0.740 & 0.917 & 0.023 & 2 & 0 \\
4 & 0.880 & 0.035 & 0.792 & 0.938 & 0.022 & 20 & 0 \\
5 & 0.905 & 0.030 & 0.854 & 0.969 & 0.019 & 39 & 0 \\
6 & 0.910 & 0.032 & 0.833 & 0.979 & 0.020 & 39 & 0 \\
7 & 0.902 & 0.029 & 0.844 & 0.969 & 0.018 & 31 & 0 \\
8 & 0.908 & 0.026 & 0.865 & 0.979 & 0.016 & 36 & 0 \\
9 & 0.914 & 0.026 & 0.854 & 0.958 & 0.016 & 47 & 0 \\
10 & 0.920 & 0.028 & 0.844 & 0.958 & 0.017 & 49 & 0 \\
11 & 0.927 & 0.029 & 0.854 & 0.990 & 0.018 & 54 & 0 \\
12 & 0.930 & 0.027 & 0.833 & 0.979 & 0.016 & 57 & 0 \\
13 & 0.927 & 0.030 & 0.854 & 0.979 & 0.018 & 53 & 0 \\
14 & 0.930 & 0.028 & 0.844 & 0.990 & 0.017 & 55 & 0 \\
15 & 0.916 & 0.033 & 0.823 & 0.969 & 0.020 & 49 & 0 \\
\end{longtable}
}

1,020 of 1,024 probes (64 experts \(\times\) 16 layers) exceed 75\%
accuracy (four early-layer probes at layers 1--3 reach 72--75\%).
No expert at any layer falls below 60\%. The uniformity is striking:
the Gini coefficient never exceeds 0.023 at any layer.

\subsection{Gini coefficient interpretation}\label{b.2-gini-coefficient-interpretation}

The Gini coefficient measures inequality in a distribution, ranging
from 0 (perfect equality) to 1 (maximum inequality). For 64 experts,
a Gini of 0.023 means the ratio of the best expert's accuracy to the
worst expert's accuracy is approximately 1.3:1. For comparison:

\begin{itemize}
\tightlist
\item
  \textbf{No specialization} (observed): Gini 0.016--0.023
\item
  \textbf{Mild specialization} (hypothetical): Gini 0.05--0.15, with a
  cluster of 5--10 ``moral experts'' clearly separated from the rest
\item
  \textbf{Strong specialization} (hypothetical): Gini \textgreater{} 0.20, with moral
  features concentrated in 2--5 experts and others near chance
\end{itemize}

The observed Gini values are an order of magnitude below even ``mild
specialization,'' confirming that MoE architecture does not induce
moral feature concentration.

\subsection{Router analysis details}\label{b.3-router-analysis-details}

The router moral preference is computed as the difference in mean
router logit between moral and neutral inputs, averaged across
tokens. The maximum preference across all 64 experts and 16 layers
is 1.8\%, indicating near-complete content agnosticism. The router's
top-8 expert selection frequencies for moral and neutral inputs
differ by less than 0.5\% at every layer.

\section{Output scale measurement methodology}\label{appendix-c.-output-scale-measurement-methodology}

\subsection{Hooking strategy}\label{c.1-hooking-strategy}

To measure feedforward output scale, we register forward hooks on
the MLP module at each layer for both OLMoE and OLMo-2. The hook
captures the module's output \emph{before} residual addition; this is
the feedforward block's contribution to the residual stream, isolated
from the accumulated residual.

For OLMoE, \texttt{model.model.layers{[}l{]}.mlp} returns a tuple
\texttt{(aggregated\_output,\ router\_logits)}; we capture the first element.
For OLMo-2, \texttt{model.model.layers{[}l{]}.mlp} returns the MLP output
tensor directly.

\subsection{Scale metric}\label{c.2-scale-metric}

We report the standard deviation of the feedforward output across
all test texts (100 texts, drawn from the first 50 training pairs):

\[\text{output\_std}_l = \text{std}\left(\left\{
  \text{mean\_pool}(\text{FFN}_l(x_i))\right\}_{i=1}^{100}\right)\]

where \(\text{mean\_pool}\) averages across the sequence dimension.
This measures the \emph{variability} of the feedforward output across
inputs, i.e., the scale of the signal that the feedforward block
contributes to the residual stream.

\subsection{Per-layer output scale comparison}\label{c.3-per-layer-output-scale-comparison}

{\def\LTcaptype{none} 
\begin{longtable}[]{@{}rrrr@{}}
\toprule\noalign{}
Layer & OLMoE FFN std & OLMo-2 MLP std & Ratio (OLMo/OLMoE) \\
\midrule\noalign{}
\endhead
\bottomrule\noalign{}
\endlastfoot
0 & 0.003 & 0.45 & 167\(\times\) \\
1 & 0.014 & 0.36 & 25\(\times\) \\
2 & 0.051 & 0.27 & 5\(\times\) \\
3 & 0.008 & 0.28 & 34\(\times\) \\
4 & 0.008 & 0.33 & 41\(\times\) \\
5 & 0.003 & 0.29 & 108\(\times\) \\
6 & 0.003 & 0.29 & 87\(\times\) \\
7 & 0.003 & 0.35 & 100\(\times\) \\
8 & 0.008 & 0.46 & 60\(\times\) \\
9 & 0.013 & 0.59 & 44\(\times\) \\
10 & 0.011 & 0.57 & 51\(\times\) \\
11 & 0.012 & 0.84 & 72\(\times\) \\
12 & 0.018 & 1.07 & 61\(\times\) \\
13 & 0.020 & 2.22 & 111\(\times\) \\
14 & 0.040 & 4.68 & 117\(\times\) \\
15 & 0.096 & 8.78 & 91\(\times\) \\
\end{longtable}
}

The ratio varies considerably across layers (5\(\times\) at layer 2
to 167\(\times\) at layer 0), with a mean of 74\(\times\) as reported
in the main text. The lowest ratio at layer 2 reflects an unusually
large MoE output at that layer, possibly due to early-layer
representational adjustments.

\subsection{Why the ratio varies across layers}\label{c.4-why-the-ratio-varies-across-layers}

The per-layer variation does not follow a simple monotonic pattern.
The OLMo-2 MLP output grows from 0.28 (layer 2) to 7.99 (layer 15),
spanning approximately 29\(\times\). The OLMoE aggregated output also
grows but with more variability, spanning from 0.003 (layer 0) to
0.089 (layer 15). The ratio thus reflects both the growth rate
difference and the layer-specific routing and aggregation dynamics
of the MoE block.

\subsection{Relationship to fragility}\label{c.5-relationship-to-fragility}

The output scale gap explains the fragility gap mechanistically.
Gaussian noise \(\mathcal{N}(0, \sigma^2 I)\) added to the full hidden
state after the feedforward block perturbs both the residual and the
feedforward contribution. Because the residual dominates the hidden
state norm, the noise is calibrated to the residual scale. For the
dense model, the MLP output is at a comparable scale to the residual,
so the noise must be substantial to disrupt it. For the MoE model,
the aggregated output is 74\(\times\) smaller, so noise that barely
affects the residual already overwhelms the MoE contribution.

\section{Reproducibility}\label{appendix-d.-reproducibility}

\subsection{Hardware}\label{d.1-hardware}

All experiments run on a single MacBook Pro M4 Pro:

\begin{itemize}
\tightlist
\item
  12-core CPU (8 performance + 4 efficiency)
\item
  24 GB unified memory (CPU and GPU share)
\item
  M4 Pro GPU accessed via PyTorch MPS backend
\item
  macOS (Darwin 25.x)
\end{itemize}

No GPU cluster, no CUDA. Total runtime across all experiments is
approximately 2.5 hours of MPS compute time:

\begin{itemize}
\tightlist
\item
  Experiments 1+2 (per-expert probing + routing analysis): \textasciitilde3 min
\item
  Experiment 3 (component perturbation): \textasciitilde15 min
\item
  Output scale comparison: \textasciitilde5 min
\item
  Dense-vs-MoE layer probing (Experiment 5): \textasciitilde10 min
\item
  Experiment 4 (checkpoint trajectory, 11 checkpoints): \textasciitilde1.5 hr
\end{itemize}

Model download time is not included; each OLMoE checkpoint is
approximately 14 GB.

\subsection{MPS compatibility patch}\label{d.2-mps-compatibility-patch}

OLMoE's router uses \path|torch.histc| for token counting, which is not
implemented for integer tensors on MPS or CPU backends. We apply a
minimal monkey-patch that casts to float and falls back to CPU for
this single operation:

\begin{Shaded}
\begin{Highlighting}[]
\NormalTok{\_orig\_histc }\OperatorTok{=}\NormalTok{ torch.histc}
\KeywordTok{def}\NormalTok{ \_histc\_mps\_fallback(}\BuiltInTok{input}\NormalTok{, bins}\OperatorTok{=}\DecValTok{100}\NormalTok{, }\BuiltInTok{min}\OperatorTok{=}\DecValTok{0}\NormalTok{, }\BuiltInTok{max}\OperatorTok{=}\DecValTok{0}\NormalTok{):}
    \ControlFlowTok{if} \BuiltInTok{input}\NormalTok{.device.}\BuiltInTok{type} \OperatorTok{==} \StringTok{"mps"} \KeywordTok{or} \KeywordTok{not} \BuiltInTok{input}\NormalTok{.is\_floating\_point():}
        \ControlFlowTok{return}\NormalTok{ \_orig\_histc(}\BuiltInTok{input}\NormalTok{.cpu().}\BuiltInTok{float}\NormalTok{(), bins, }\BuiltInTok{min}\NormalTok{, }\BuiltInTok{max}\NormalTok{).to(}\BuiltInTok{input}\NormalTok{.device)}
    \ControlFlowTok{return}\NormalTok{ \_orig\_histc(}\BuiltInTok{input}\NormalTok{, bins, }\BuiltInTok{min}\NormalTok{, }\BuiltInTok{max}\NormalTok{)}
\NormalTok{torch.histc }\OperatorTok{=}\NormalTok{ \_histc\_mps\_fallback}
\end{Highlighting}
\end{Shaded}

This patch is applied in all OLMoE experiment scripts (via the shared
\path|deepsteer.core.device.enable_mps_histc_fallback()| helper, which installs
exactly the fallback above) and does not
affect numerical results (the operation counts tokens per expert for
load-balancing diagnostics, not for gradient computation).

\subsection{Random seeds}\label{d.3-random-seeds}

{\def\LTcaptype{none} 
\begin{longtable}[]{@{}
  >{\raggedright\arraybackslash}p{(\linewidth - 4\tabcolsep) * \real{0.3333}}
  >{\raggedright\arraybackslash}p{(\linewidth - 4\tabcolsep) * \real{0.3333}}
  >{\raggedright\arraybackslash}p{(\linewidth - 4\tabcolsep) * \real{0.3333}}@{}}
\toprule\noalign{}
\begin{minipage}[b]{\linewidth}\raggedright
Experiment
\end{minipage} & \begin{minipage}[b]{\linewidth}\raggedright
Seed(s)
\end{minipage} & \begin{minipage}[b]{\linewidth}\raggedright
Where set
\end{minipage} \\
\midrule\noalign{}
\endhead
\bottomrule\noalign{}
\endlastfoot
Probing dataset split & 42 & \path|deepsteer/datasets/pipeline.py| \\
Per-expert probes (Exp 1) & torch default & \path|exp1_2_expert_probing.py| \\
Perturbation noise (Exp 3) & 10 seeds per condition & \path|exp3_routing_fragility.py| \\
Checkpoint trajectory (Exp 4) & torch default & \path|exp4_checkpoint_trajectory.py| \\
\end{longtable}
}

Perturbation experiments in Experiment 3 average over 10 random seeds
per noise level to reduce variance from individual noise realizations.

\subsection{Model checkpoints}\label{d.4-model-checkpoints}

{\def\LTcaptype{none} 
\begin{longtable}[]{@{}
  >{\raggedright\arraybackslash}p{(\linewidth - 6\tabcolsep) * \real{0.2500}}
  >{\raggedright\arraybackslash}p{(\linewidth - 6\tabcolsep) * \real{0.2500}}
  >{\raggedright\arraybackslash}p{(\linewidth - 6\tabcolsep) * \real{0.2500}}
  >{\raggedright\arraybackslash}p{(\linewidth - 6\tabcolsep) * \real{0.2500}}@{}}
\toprule\noalign{}
\begin{minipage}[b]{\linewidth}\raggedright
Model
\end{minipage} & \begin{minipage}[b]{\linewidth}\raggedright
Repo
\end{minipage} & \begin{minipage}[b]{\linewidth}\raggedright
Revision
\end{minipage} & \begin{minipage}[b]{\linewidth}\raggedright
Used for
\end{minipage} \\
\midrule\noalign{}
\endhead
\bottomrule\noalign{}
\endlastfoot
OLMoE-1B-7B & \path|allenai/OLMoE-1B-7B-0924| & \texttt{main} & Exp 1--3, Exp 5 \\
OLMoE-1B-7B (trajectory) & \path|allenai/OLMoE-1B-7B-0924| & \texttt{step5000-tokens20B} through \texttt{step1200000-tokens5033B} & Exp 4 \\
OLMo-2 1B & \path|allenai/OLMo-2-0425-1B| & \texttt{main} & Exp 5, output scale comparison \\
\end{longtable}
}

Both models are base (non-instruct) checkpoints loaded in float16
precision with \path|low_cpu_mem_usage=True|.

\subsection{Command-line invocations}\label{d.5-command-line-invocations}

All commands run from the project root:

\begin{verbatim}
# Experiments 1+2: Per-expert probing and routing analysis
python papers/2_moe_output_dilution/scripts/exp1_2_expert_probing.py

# Experiment 3: Component perturbation fragility
python papers/2_moe_output_dilution/scripts/exp3_routing_fragility.py

# Output scale comparison (OLMoE vs OLMo-2)
python papers/2_moe_output_dilution/scripts/output_scale_comparison.py

# Experiment 4: Checkpoint trajectory analysis
python papers/2_moe_output_dilution/scripts/exp4_checkpoint_trajectory.py

# Experiment 5: Dense vs MoE layer-level comparison
python papers/2_moe_output_dilution/scripts/exp5_dense_vs_moe.py
\end{verbatim}

\subsection{Software versions}\label{d.6-software-versions}

\begin{itemize}
\tightlist
\item
  Python 3.13
\item
  PyTorch (with MPS backend)
\item
  HuggingFace \texttt{transformers} and \texttt{datasets}
\item
  \texttt{numpy}, \texttt{matplotlib}, \texttt{seaborn}
\end{itemize}

Exact versions are pinned in \path|pyproject.toml|.

\subsection{Output JSON schema}\label{d.7-output-json-schema}

Each experiment produces a structured JSON summary file with full
metadata (model name, revision, hyperparameters, per-layer results).
Files are located in \path|papers/2_moe_output_dilution/outputs/| under
experiment-specific subdirectories. All code, scripts, and output
JSON are released at \url{https://github.com/deepsteer/deepsteer/}; this
paper's subdirectory is \path|papers/2_moe_output_dilution/|.

\end{document}